\pdfoutput=1

\documentclass[11pt]{article}

\usepackage[]{acl}

\usepackage{times}
\usepackage{latexsym}
\usepackage{amsmath}
\usepackage{amsfonts}
\usepackage{booktabs}
\usepackage{graphicx}
\usepackage{inconsolata}
\usepackage{bbold}
\usepackage{colortbl}

\usepackage[T1]{fontenc}

\usepackage[utf8]{inputenc}

\usepackage{microtype}

\title{Diagnosing Transformers in Task-Oriented Semantic Parsing}

\author{Shrey Desai \quad\quad Ahmed Aly \\
  Facebook \\
  \texttt{\{shreyd, ahhegazy\}@fb.com}}

\begin{document}
\maketitle
\begin{abstract}
Modern task-oriented semantic parsing approaches typically use seq2seq transformers to map textual utterances to semantic frames comprised of intents and slots. While these models are empirically strong, their specific strengths and weaknesses have largely remained unexplored. In this work, we study BART \cite{lewis-2020-bart} and XLM-R \cite{conneau-2020-xlmr}, two state-of-the-art parsers, across both monolingual and multilingual settings. Our experiments yield several key results: transformer-based parsers struggle not only with disambiguating intents/slots, but surprisingly also with producing syntactically-valid frames. Though pre-training imbues transformers with syntactic inductive biases, we find the ambiguity of copying utterance spans into frames often leads to tree invalidity, indicating span extraction is a major bottleneck for current parsers. However, as a silver lining, we show transformer-based parsers give sufficient indicators for whether a frame is likely to be correct or incorrect, making them easier to deploy in production settings.
\end{abstract}

\section{Introduction}

Task-oriented semantic parsing---mapping textual utterances to semantic frames---is a critical component of modern conversational AI systems \cite{gupta-2020-top,aghajanyan-2020-decoupled}. Recent methodology casts parsing as transduction, using seq2seq pre-trained transformers to produce linearized parse trees \cite{aghajanyan-2020-decoupled,chen-2020-topv2,li-2020-mtop}; here, each frame token is either \textit{copied} from the utterance or \textit{generated} from an ontology. Compared to explicit grammar-based approaches \cite{gupta-2020-top}, this plug-and-play of transformers simplifies the learning objective and scales to multilingual settings, but the lack of provenance makes it challenging to understand model behavior ``under the hood.''

\begin{figure}[t]
    \centering
    \includegraphics[scale=0.25]{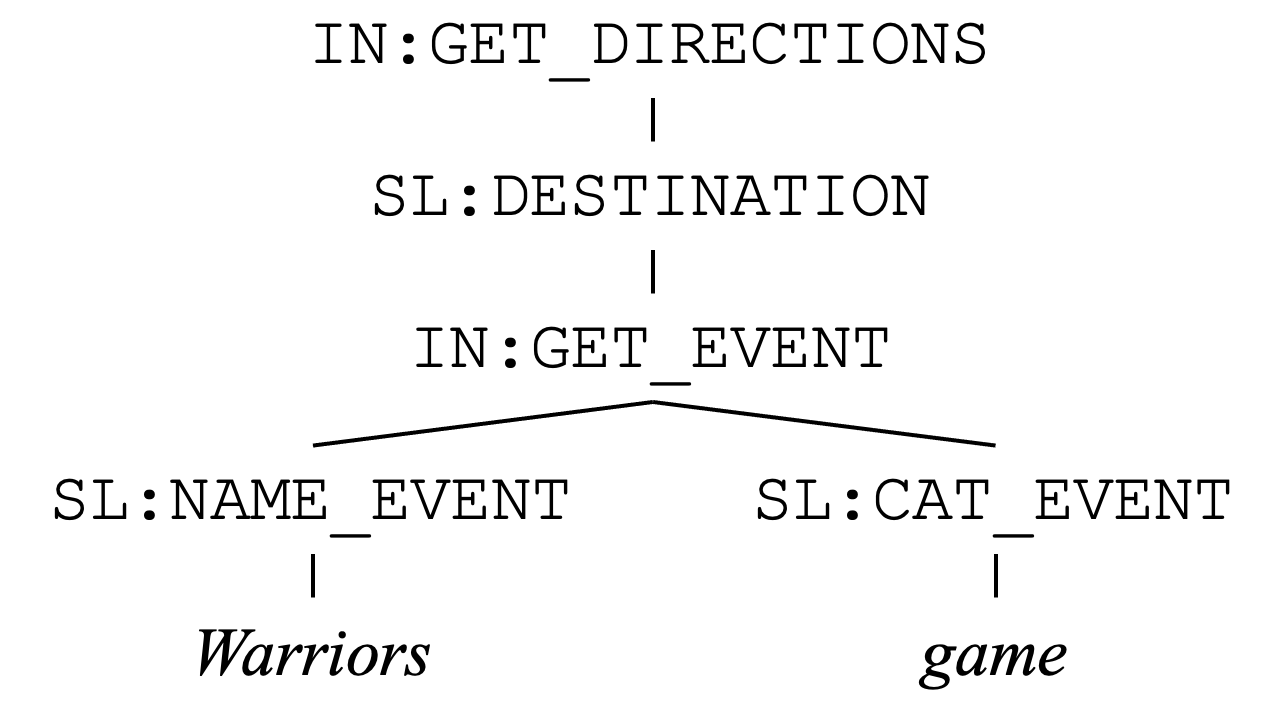}
    \caption{Example decoupled semantic frame representation \cite{aghajanyan-2020-decoupled} for the utterance \textit{Directions to the Warriors game}.}
    \label{fig:decoupled}
\end{figure}

In this work, we investigate the strengths and weaknesses of transformer-based semantic parsers and provide modeling directions based on data-driven insights. Specifically, we study BART \cite{lewis-2020-bart} and XLM-R \cite{conneau-2020-xlmr}, two state-of-the-art conversational semantic parsers, on both monolingual (TOP/TOPv2; \cite{gupta-2020-top,chen-2020-topv2}) and multilingual (MTOP; \cite{li-2020-mtop}) datasets. The compositionality of utterances in these datasets provide a strong testbed for resolving both complex syntactic structure and semantic ambiguity, mirroring the types of challenges our parsers are likely to encounter in practice.

We design our experiments around three main questions. First, broadly speaking, what types of errors do transformer-based parsers make? We begin by annotating 500+ predicted frames across 6 languages and categorize them with fine-grained types. We find transformer-based parsers struggle not only with \textit{classification} (i.e., disambiguating intents/slots) but also \textit{planning} (i.e., switching between copying/generating). Planning errors are more egregious: misplacing close brackets, for example, can violate tree constraints, rendering the entire frame unusable.

Next, we investigate transformer-based parsers' abilities to generate syntactically-valid trees. Specifically, are planning mistakes caused by general uncertainty, or worse, a pathology of seq2seq learning? To address this, we devise an oracle setting where a model conditions on partially gold information (either utterance spans or syntactic structure) and predicts the remaining parts of the frame. Surprisingly, we find conditioning on gold spans---not gold structures---results in near-perfect trees at most depths, pointing towards span extraction as a major bottleneck for current parsers.

Finally, though transformer-based parsers are susceptible to error, ideally, we should be able to proactively diagnose mistakes. Using features from model generations (e.g., confidence), can we intrinsically judge if a sequence is correct or incorrect? Encouragingly, we show that a confidence estimation system combining a transformer-based parser and feature-based classifier can detect \textit{correct} frames with 90\%+ F1, indicating usability in production settings.

\section{Experimental Setup}

\begin{table}[t]
\centering
\small
\setlength{\tabcolsep}{4pt}
\begin{tabular}{lrrr}
\toprule
split & TOP & TOPv2 & MTOP \\
\midrule
train & 31,279 & 124,579 & 73,956 \\
dev & 4,462 & 17,160 & 10,852 \\
test & 9,042 & 38,785 & 30,541 \\
\bottomrule
\end{tabular}
\caption{Dataset splits for TOP, TOPv2, and MTOP.}
\label{tab:dataset-splits}
\end{table}

\begin{table}[t]
\centering
\small
\setlength{\tabcolsep}{4pt}
\begin{tabular}{rrrrrr}
\toprule
$N$ & $d_\textrm{model}$ & $d_\textrm{ff}$ & $h$ & $d_k$ & $d_v$ \\
\midrule
6 & 1024 & 4096 & 16 & 64 & 64 \\
\bottomrule
\end{tabular}
\caption{Dimensions of transformer decoder added to XLM-R for MTOP fine-tuning. Notation is borrowed from \citet{vaswani-2017-transformer}.}
\label{tab:decoder-dim}
\end{table}

\begin{table}[t]
\centering
\small
\setlength{\tabcolsep}{4pt}
\begin{tabular}{lrrr}
\toprule
split & TOP & TOPv2 & MTOP \\
\midrule
dev & 85.41 & 87.53 & 76.00 \\
test & 85.74 & 87.52 & 77.20 \\
\bottomrule
\end{tabular}
\caption{Exact match (EM) of BART and XLM-R on TOP/TOPv2 and MTOP, respectively.}
\label{tab:dataset-perf}
\end{table}

\begin{table}[t]
\centering
\small
\setlength{\tabcolsep}{4pt}
\begin{tabular}{lrrr}
\toprule
setting & TOP & TOPv2 & MTOP \\
\midrule
model & BART & BART & XLM-R \\
dropout & 8.68e-2 & 1.82e-1 & 0 \\
batch size & 16 & 16 & 16 \\
epochs & 50 & 50 & 50 \\
optimizer & Lamb & Lamb & Lamb \\
lr & 3.72e-4 & 4.88e-4 & 6.91e-4 \\
weight decay & 6.25e-7 & 6.26e-7 & 6.25e-7 \\
swa lr & 2.08e-4 & 1.86e-4 & 3.96e-4 \\
swa start & 8945 & 18876 & 19450 \\
swa freq & 219 & 233 & 185 \\
scheduler & exp & exp & exp \\
warmup & 5000 & 5000 & 5000 \\
gamma & 0.95 & 0.95 & 0.95 \\
\bottomrule
\end{tabular}
\caption{Hyperparameters for fine-tuning models on TOP, TOPv2, and MTOP.}
\label{tab:hyperparams}
\end{table}

We conduct experiments on the following task-oriented semantic parsing datasets: (1) \textbf{TOP:} parallel corpus consisting of English utterances and corresponding semantic frames \cite{gupta-2020-top}; (2) \textbf{TOPv2:} monolingual extension of TOP to 6 domains \cite{chen-2020-topv2}; (3) \textbf{MTOP:} multilingual extension of TOP spanning English, Spanish, French, German, Hindi, and Thai \cite{li-2020-mtop}. Table~\ref{tab:dataset-splits} shows train, dev, and test splits for the datasets.

Each dataset sample consists of a textual utterance $x$ and (linearized) semantic frame $y$. Here, frames are in decoupled form \cite{aghajanyan-2020-decoupled}, as each token is derived either from \textit{copying} from the utterance or \textit{generating} from the ontology (see Figure~\ref{fig:decoupled}). Following prior work, we fine-tune seq2seq transformers to maximize the log likelihood of the gold frame token at each timestep: $\sum_{(x, y)} \sum_t \log P(y_t | y_{<t}, x;\theta)$.

On TOP/TOPv2, we fine-tune BART \cite{lewis-2020-bart}, a seq2seq transformer pre-trained with a denoising autoencoder objective on monolingual corpora, and on MTOP, we fine-tune XLM-R \cite{conneau-2020-xlmr} (equipped with a randomly-initialized decoder), a transformer encoder pre-trained with a masked language modeling objective on multilingual corpora. For XLM-R, specifically, we attach a randomly-initialized decoder (see Table~\ref{tab:decoder-dim}). Table~\ref{tab:dataset-perf} shows model performance as judged by exact match. Hyperparameters for all models are listed in Table~\ref{tab:hyperparams}.

\section{Error Analysis}

In this section, we seek to better understand the types of errors transformer-based parsers make across both monolingual and multilingual settings. 

\begin{table}[t]
\small
\setlength{\tabcolsep}{4pt}
\begin{tabular}{lrrrrrr}
\toprule
 & \multicolumn{3}{c}{Exact Match} & \multicolumn{3}{c}{Tree Validity} \\
 \cmidrule(lr){2-4} \cmidrule(lr){5-7}
$d$ & \quad TOP & TOPv2 & MTOP & \quad TOP & TOPv2 & MTOP \\
\midrule
1 & \cellcolor{red!24.41}{78.03} & \cellcolor{red!14.91}{86.58} & \cellcolor{red!16.94}{84.75} & \cellcolor{red!1.50}{98.65} & \cellcolor{red!6.03}{94.57} & \cellcolor{red!9.74}{91.23} \\
2 & \cellcolor{red!8.56}{92.30} & \cellcolor{red!10.37}{90.67} & \cellcolor{red!15.86}{85.73} & \cellcolor{red!3.37}{96.97} & \cellcolor{red!3.53}{96.82} & \cellcolor{red!6.89}{93.80} \\
3 & \cellcolor{red!10.07}{90.94} & \cellcolor{red!12.78}{88.50} & \cellcolor{red!28.27}{74.56} & \cellcolor{red!3.22}{97.10} & \cellcolor{red!4.06}{96.35} & \cellcolor{red!10.17}{90.85} \\
4 & \cellcolor{red!13.07}{88.24} & \cellcolor{red!15.20}{86.32} & \cellcolor{red!39.41}{64.53} & \cellcolor{red!4.52}{95.93} & \cellcolor{red!5.03}{95.47} & \cellcolor{red!15.86}{85.73} \\
5 & \cellcolor{red!18.46}{83.39} & \cellcolor{red!18.19}{83.63} & \cellcolor{red!61.90}{44.29} & \cellcolor{red!6.34}{94.29} & \cellcolor{red!5.72}{94.85} & \cellcolor{red!33.83}{69.55} \\
6 & \cellcolor{red!18.82}{83.06} & \cellcolor{red!17.18}{84.54} & \cellcolor{red!61.73}{44.44} & \cellcolor{red!6.67}{94.00} & \cellcolor{red!6.17}{94.45} & \cellcolor{red!41.67}{62.50} \\
\bottomrule
\end{tabular}
\caption{Benchmarks of BART and XLM-R on TOP/TOPv2 and MTOP, respectively, according to exact match and tree validity at increasing tree depths ($d$).}
\label{tab:model-perf}
\end{table}

\subsection{Error Types}
\label{sec:error-types}

To standardize our analysis, we categorize model errors under the following types: \textbf{intent} (incorrect intent prediction), \textbf{slot} (incorrect slot prediction), \textbf{out-of-domain} (incorrect out-of-domain intent prediction), \textbf{mode} (confusion between copying an utterance token or generating an ontology token), and \textbf{leaf} (incorrect span in a frame leaf slot). In addition, we report the syntactic \textbf{validity} of parse trees separately, though we note mode errors typically result in invalid constructions.

One complicating factor is that a predicted sequence may potentially contain several errors, and because decoding is conducted autoregressively, a given error may be influenced by earlier errors (if any such exist). Therefore, to reduce the number of confounding variables, we only consider settings where an incorrect prediction has gold history $\mathrm{argmax}_{y_i} P(y_i|y^{*}_{<t},x) \ne y^{*}_i$; put another way, we only count the \textit{first} error in a sequence.

Using the framework discussed above, we annotate 700 errors across BART and XLM-R on TOP and MTOP, respectively; 100 errors are from TOP and 6$\times$100 errors are from MTOP (100 per language).

\begin{figure}[t]
    \centering
    \includegraphics[scale=0.5]{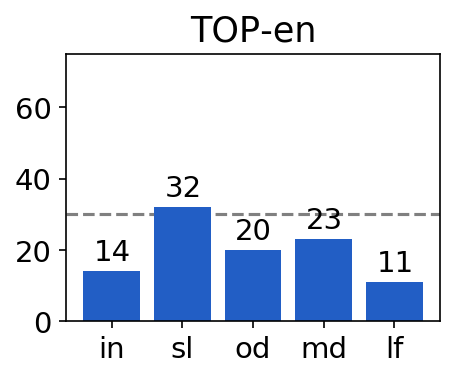}
    \includegraphics[scale=0.5]{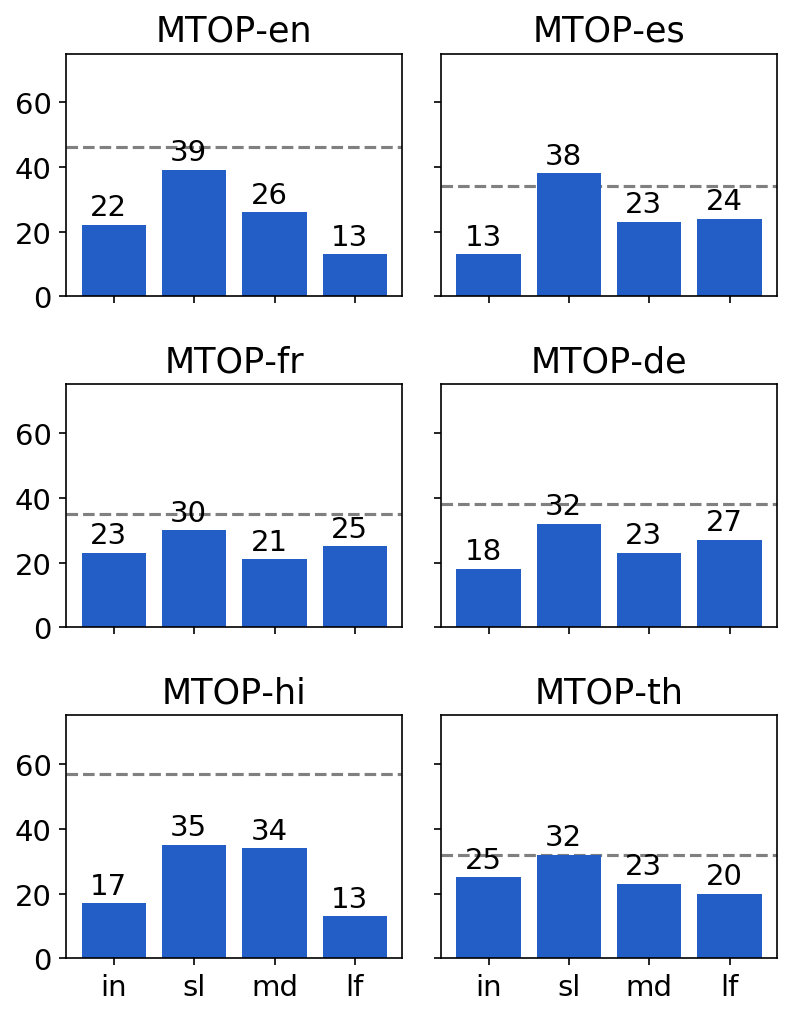}
    \caption{Distribution of errors across TOP and MTOP categorized by intent (in), slot (sl), out-of-domain (od), mode (md), and leaf (lf). Dashed lines indicate the percentage of trees which are syntactically valid.}
    \label{fig:error-analysis}
\end{figure}

\subsection{Results}

Table~\ref{tab:model-perf} benchmarks overall model performance and Figure~\ref{fig:error-analysis} categorizes errors with fine-grained types; from these results, we draw the following conclusions:

\paragraph{Transformer-based parsers typically struggle with both classification and planning.} In the seq2seq formulation, models must jointly \textit{classify} (i.e., provide intent and slot labels) and \textit{plan} (i.e., switch between copying and generating) when producing a semantic frame. Our results show intent/slot and mode errors, which generally fall under the theme of classification and planning, respectively, account for nearly 70-80\% of errors. A key observation, however, is that classification and planning error statistics are relatively consistent across languages, suggesting our models may not need language-specific fine-tuning to address these particular errors.

\paragraph{Nearly 40\% of incorrectly predicted frames are syntactically invalid.} Surprisingly, a large percentage of incorrectly predicted frames violate tree constraints; for linearized frames, this implies the number of open brackets (\texttt{[in} or \texttt{[sl}) do not match the number of close brackets (\texttt{]}). Though well-formedness \textit{is} correlated with depth, we see tree validity (1) is not substantially improved by increasing the number of monolingual samples (TOP $\rightarrow$ TOPv2) and (2) drops off quite rapidly for multilingual samples (TOP/TOPv2 $\rightarrow$ MTOP).

\paragraph{Span extraction is more challenging in multilingual settings.} Leaf errors in English \texttt{(TOP|MTOP)-en} are typically twice as lower compared to those in non-English languages \texttt{MTOP-(es|fr|de|hi|th)}. Upon closer inspection, we find most leaf errors in English are relatively benign; the model may drop a preposition when copying a span (e.g., \textit{Monday} as opposed to \textit{on Monday}). However, for languages beyond English, extracted spans in leaf slots typically consist of hallucinated or duplicated subwords, which are much more serious in nature. Finally, though languages with non-projective structures (e.g., German) can populate leaf slots with non-contiguous spans, we noticed errors on these types of samples were infrequent.

\paragraph{Out-of-domain detection is also a significant source of error.} TOP, in particular, mixes the canonical semantic parsing task with out-of-domain detection by assigning such utterances the frame \texttt{[in:unsupported ]}.\footnote{There also exist more fine-grained out-of-domain categories, such as \texttt{[in:unsupported-event ]}.} Though well-motivated, roughly 20\% of errors are related to incorrect out-of-domain predictions, suggesting our models have not precisely learned the boundary between in-domain and out-of-domain utterances. If high detection accuracy is preferred, multi-tasking parsers in this fashion may not be an effective use of parameters (assuming more data is not available); instead, out-of-domain detection can be conducted independently with alternate methodology \cite{gangal-2019-ood}.

\section{Syntactic Structure}

Our case study above demonstrates transformer-based parsers can produce syntactically-invalid frames at a high rate. These structural errors are more serious than disambiguation errors since they render the frame unusable, potentially causing cascading failures in a task-oriented dialog system. Therefore, in this section, we dive deeper into why tree constraints are not satisfied and question the possibility of achieving perfect tree validity.

While transduction models do not explicitly impose tree constraints, there is precedent that strong neural representations do implicitly model tree structures; recent studies demonstrate large-scale pre-training, in particular, imbues strong notions of syntax \cite{goldberg-2019-bert,jawahar-2019-bert,tenney-2019-bert}. Taking these results together, we hypothesize that transformer representations may be ``good enough'', but instead there exist ambiguous aspects of our task-oriented semantic parsing task which cause tree invalidity.

Previously, we saw transformer-based semantic parsers largely struggled with classification- and planning-related errors. Therefore, the question we pose is: if we resolve these ambiguities by creating oracle models, can we achieve perfect tree validity? This setup also enables us to gain a deeper understand of the upper-bound performance of transformer-based semantic parsers, even as their representations get stronger.

\begin{table*}[t]
\centering
\small
\begin{tabular}{lll}
\toprule
model type & utterance $x$ (+ snippet $z$) & frame $y$ \\
\midrule
regular & Where can I see fireworks tonight? & \texttt{[in [sl} fireworks \texttt{[sl} tonight \texttt{] ]} \\
span oracle & \quad+ \texttt{[span1]} fireworks \texttt{[span2]} tonight &  \\
struct oracle & \quad+ \texttt{[in [sl [span1] [sl [span2] ] ]} & \\
\bottomrule
\end{tabular}
\caption{Example source and target pairs for oracle experiments. The span oracle specifies the gold spans while the struct oracle specifies the gold structure. Note that \texttt{[in} and \texttt{[sl} are used for brevity.}
\label{tab:oracle-exs}
\end{table*}

\paragraph{Oracle Models.} Because classification and planning target inherently different phenomena, creating an oracle that simultaneously makes both less ambiguous is challenging. Instead, we experiment with two separate oracles---\textbf{span oracle} and \textbf{structure oracle} models for classification and planning, respectively---which map an utterance $x$ along with a ``partially gold'' snippet $z$ to generate the frame $y$, inducing the objective $\sum_{(x, y, z)} \sum_t \log P(y_t | y_{<t}, x, z;\theta)$. 

For example, given an utterance $x$ \textit{Where can I see fireworks tonight?} and frame $y$ \texttt{[in [sl} \textit{fireworks} \texttt{[sl} \textit{tonight} \texttt{] ]}, the span oracle model defines $z$ as \texttt{[span1]} \textit{fireworks} \texttt{[span2]} \textit{tonight} and the structure oracle model defines $z$ as \texttt{[in [sl [span1] [sl [span2] ] ]}.\footnote{Fine-grained intent/slot labels are omitted for visual clarity, but are included during model training.} Here, providing $z$ as input helps the model learn $y \setminus z$; span oracle models optimize for correct structure and structure oracle models optimize for correct spans. Table~\ref{tab:oracle-exs} shows example source and target pairs for the regular, span oracle, and structure oracle models.

\begin{figure}[t]
    \centering
    \includegraphics[scale=0.55]{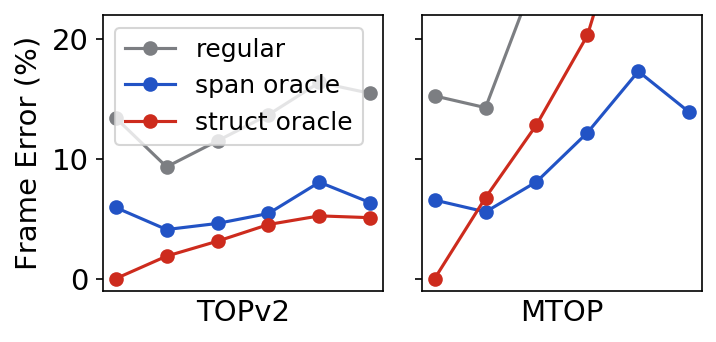}
    \includegraphics[scale=0.55]{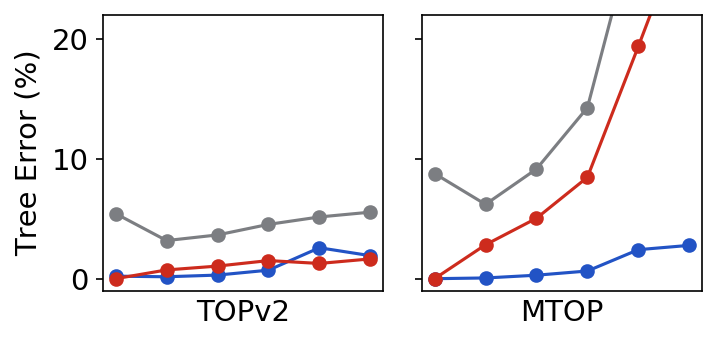}
    \caption{Exact match (EM) and tree validity (TV) error (\%) of the regular, span oracle, and structure oracle models on TOPv2 and MTOP. Dots from left $\rightarrow$ right indicate increasing frame compositionality (the graph depths of 1 $\rightarrow$ 6).}
    \label{fig:synthetic-results}
\end{figure}

\paragraph{Results.} Figure~\ref{fig:synthetic-results} shows the oracle model results; we measure both exact match and tree validity error. \textbf{A key phenomenon we observe is that conditioning on gold spans results in near-zero tree validity error at most depths.} Surprisingly, we see conditioning on gold structures (to stress, the \textit{exact} syntactic structure) never consistently results in well-formed trees, especially as the depth increases. Structure oracle models still suffer from mode errors during generation: augmenting a leaf span with an extra word instead of placing a close bracket, for example, is a typical mistake. Furthermore, we see this problem is magnified in MTOP, which connects to the notion that span extraction tends to be difficult in multilingual settings.

Our experiments suggest seq2seq transformer-based parsers \textit{can} achieve near-perfect tree validity---even at large depths---provided that span extraction is precise. Currently, however, this is a major source of ambiguity our parsers are not well-equipped to handle, especially when scaling to languages beyond English.

\section{Confidence Estimation}

Despite the criticism we have presented of state-of-the-art, transformer-based conversational semantic parsers, these models do demonstrate strong performance over prior baselines, and correctly parse a vast majority of samples. A property that can make these models easier to deploy in practice is if they ``know what they don't know'' \cite{desai-2020-calibration}; besides interpretability, this is particularly useful for identifying and correcting errors in tail scenarios via active learning \cite{dredze-2008-al,duong-2018-al,sen-2020-al}. We frame this problem as confidence estimation \cite{blatz-2004-ce}: given an utterance $x$, predicted frame $y'$, and gold frame $y$, we seek to learn a binary classifier which uses target-side features $f(y')$ to estimate $P(y'=y) = \mathrm{sigmoid}(w^\top f(y'))$.

To make our approach as generalizable as possible, we constrain $f(y')$ to be as model-agnostic and recall-oriented as possible. We select the following features: (1) \textbf{length:} $|y'|$; (2) \textbf{validity:} $\max(0, \sum_i \mathbb{1}[y'_i \in V^+] - \mathbb{1}[y'_i \in V^+])$ where $V^+$ and $V^-$ are the set of open and close brackets, respectively; and (3) \textbf{confidence:} $\frac{1}{|y'|}\sum_i P(y'_t|y'_{<t},x)$. Using our best transformer-based parsers, we obtain predictions on a held-out set $D_\textrm{dev}$ and test set $D_\textrm{test}$. Then, we train and test a SVM on $D_\textrm{dev}$ and $D_\textrm{test}$, respectively, using the features defined above. 

In addition to the standard hinge loss, we also add a class imbalance penalty as positive examples are typically 5-8$\times$ as prevalent depending on the dataset. We chiefly evaluate the binary classifier's ability to identify semantic frames which are correct (i.e., the positive class). From an active learning standpoint, getting positive samples wrong is more serious than getting negative samples wrong; annotation resources are best directed towards boundary or incorrect predictions.

\begin{table}[]
\centering
\small
\setlength{\tabcolsep}{4pt}
\begin{tabular}{lrrrrrrrrr}
\toprule
 & \multicolumn{3}{c}{TOPv2} & \multicolumn{3}{c}{MTOP} \\
 \cmidrule(lr){2-4} \cmidrule(lr){5-7}
 & P & R & F1 & P & R & F1\\
\midrule
SVM & 97.2 & 85.7 & 91.2 & 95.0 & 85.2 & 89.8  \\
~~--length & \textbf{97.7} & 84.8 & 90.8 & \textbf{95.1} & 84.7 & 89.6 \\
~~--validity & 97.0 & 82.6 & 89.2 & 94.9 & 80.5 & 87.1 \\
~~--confidence & 91.6 & \textbf{98.8} & \textbf{95.1} & 85.3 & \textbf{95.8} & \textbf{90.2} \\
\bottomrule
\end{tabular}
\caption{Precision (P), recall (R), and F1 of the SVM-based confidence estimator. --$x$ indicates an ablation of feature $x$ (i.e., it is omitted during learning).}
\label{tab:ce-results}
\end{table}

Table~\ref{tab:ce-results} shows the performance and ablations of our confidence estimator. \textbf{In both monolingual and multilingual settings, using transformer-based features, we can detect \textit{correct} semantic frames with 90\%+ F1.} In particular, we see length and validity largely capture the space of correct frames (recall) and confidence effectively distinguishes between correct and incorrect frames (precision). Practitioners may select an SVM variant depending on whether precision or recall is preferred.

\section{Conclusion}

In this work, we assess the strengths and weaknesses of seq2seq transformers for task-oriented semantic parsing. These models ``know what they don't know'', making them easier to depoy in practice, but cannot perfectly model compositional utterances, as indicated by the challenges of span extraction. We believe that modeling efforts in this direction---as opposed to simply annotating more data---can improve parsers substantially.

\bibliography{custom}
\bibliographystyle{acl_natbib}

\end{document}